\documentclass[sigconf,11pt]{acmart}

\usepackage{booktabs} % For formal tables

\setcopyright{none}
\usepackage{amsmath}
\usepackage{amssymb}
\usepackage{amsthm}

\usepackage{graphicx}
\usepackage{array}
\usepackage{multirow}
\usepackage[font={small}]{caption}

\begin{document}
\title{Interpretability via Model Extraction}
\titlenote{Presented as a poster at the 2017 Workshop on Fairness, Accountability, and Transparency in Machine Learning (FAT/ML 2017).}
%  copyright information}
%\subtitle{Extended Abstract}
%\subtitlenote{The full version of the author's guide is available as
%  \texttt{acmart.pdf} document}

\author{Osbert Bastani}
%\authornote{Dr.~Trovato insisted his name be first.}
%\orcid{1234-5678-9012}
\affiliation{%
  \institution{Stanford University}}
%  \streetaddress{P.O. Box 1212}
%  \city{Dublin} 
%  \state{Ohio} 
%  \postcode{43017-6221}
%\country{USA}}
\email{obastani@cs.stanford.edu}

\author{Carolyn Kim}
%\authornote{The secretary disavows any knowledge of this author's actions.}
\affiliation{%
  \institution{Stanford University}}
%  \streetaddress{P.O. Box 1212}
%  \city{Dublin} 
%  \state{Ohio} 
%  \postcode{43017-6221}
%\country{USA}}
\email{ckim@cs.stanford.edu}

\author{Hamsa Bastani}
%\authornote{This author is the
%  one who did all the really hard work.}
\affiliation{%
  \institution{Stanford University}}
%  \streetaddress{1 Th{\o}rv{\"a}ld Circle}
%  \city{Hekla} 
%\country{USA}}

\email{hsridhar@stanford.edu}

%\author{Lawrence P. Leipuner}
%\affiliation{
%  \institution{Brookhaven Laboratories}
%  \streetaddress{P.O. Box 5000}}
%\email{lleipuner@researchlabs.org}

%\author{Sean Fogarty}
%\affiliation{%
%  \institution{NASA Ames Research Center}
%  \city{Moffett Field}
%  \state{California} 
%  \postcode{94035}}
%\email{fogartys@amesres.org}

%\author{Charles Palmer}
%\affiliation{%
%  \institution{Palmer Research Laboratories}
%  \streetaddress{8600 Datapoint Drive}
%  \city{San Antonio}
%  \state{Texas} 
%  \postcode{78229}}
%\email{cpalmer@prl.com}

%\author{John Smith}
%\affiliation{\institution{The Th{\o}rv{\"a}ld Group}}
%\email{jsmith@affiliation.org}

%\author{Julius P.~Kumquat}
%\affiliation{\institution{The Kumquat Consortium}}
%\email{jpkumquat@consortium.net}

% The default list of authors is too long for headers}
%\renewcommand{\shortauthors}{B. Trovato et al.}

\begin{abstract}
The ability to interpret machine learning models has become increasingly important now that machine learning is used to inform consequential decisions. We propose an approach called \emph{model extraction} for interpreting complex, blackbox models. Our approach approximates the complex model using a much more interpretable model; as long as the approximation quality is good, then statistical properties of the complex model are reflected in the interpretable model. We show how model extraction can be used to understand and debug random forests and neural nets trained on several datasets from the UCI Machine Learning Repository, as well as control policies learned for several classical reinforcement learning problems.
\end{abstract}

%
% The code below should be generated by the tool at
% http://dl.acm.org/ccs.cfm
% Please copy and paste the code instead of the example below. 
%
%\begin{CCSXML}
%<ccs2012>
% <concept>
%  <concept_id>10010520.10010553.10010562</concept_id>
%  <concept_desc>Computer systems organization~Embedded systems</concept_desc>
%  <concept_significance>500</concept_significance>
% </concept>
% <concept>
%  <concept_id>10010520.10010575.10010755</concept_id>
%  <concept_desc>Computer systems organization~Redundancy</concept_desc>
%  <concept_significance>300</concept_significance>
% </concept>
% <concept>
%  <concept_id>10010520.10010553.10010554</concept_id>
%  <concept_desc>Computer systems organization~Robotics</concept_desc>
%  <concept_significance>100</concept_significance>
% </concept>
% <concept>
%  <concept_id>10003033.10003083.10003095</concept_id>
%  <concept_desc>Networks~Network reliability</concept_desc>
%  <concept_significance>100</concept_significance>
% </concept>
%</ccs2012>  
%\end{CCSXML}

%\ccsdesc[500]{Computer systems organization~Embedded systems}
%\ccsdesc[300]{Computer systems organization~Redundancy}
%\ccsdesc{Computer systems organization~Robotics}
%\ccsdesc[100]{Networks~Network reliability}

%\keywords{ACM proceedings, \LaTeX, text tagging}

\maketitle

\section{Introduction}

Recent advances in machine learning have revolutionized our ability to use data to inform critical decisions, such as medical diagnosis~\cite{diagnosis2001, caruana2015intelligible, valdes2016mediboost}, bail decisions for defendants~\cite{bail2017, jung2017}, and aircraft collision avoidance systems~\cite{collision2010}. At the same time, machine learning algorithms have been shown to exhibit unexpected defects when deployed in the real world; examples include causality (i.e., inability to distinguish causal effects from correlations)~\cite{pearl2009, caruana2015intelligible}, fairness (i.e., internalizing prejudices present in training data)~\cite{dwork2012fairness, fairness2016},
% covariate shift (i.e., differences in the training and test distributions)~\cite{covariateshift2000, covariateshift2009},
and algorithm aversion (i.e., lack of trust by end users)~\cite{dietvorst2015}.

Interpretability is a promising approach to address these challenges~\cite{rudin2014algorithms,doshi2017roadmap}---we can help human users diagnose issues and verify correctness of machine learning models by providing insight into the model's reasoning~\cite{slim,rulelists,lime,letham2015interpretable,koh2017understanding}. For example, suppose the user is trying to train a model that does not depend on a prejudiced feature (e.g., ethnicity). Omitting the feature might not suffice to avoid prejudice, since the model could reconstruct that feature from other features~\cite{pedreshi2008discrimination}. A better approach might be to train the model with the prejudiced feature, and then assess the dependence of the model on that feature. This approach requires the ability to understand the model's reasoning process, i.e., how model predictions are affected by changing the prejudiced feature~\cite{doshi2017roadmap}. Similarly, the user may want to determine whether dependence on a feature is causal, or understand the high-level structure of the model to gain confidence in its correctness.

In this paper, we propose an technique that we call \emph{model extraction} for interpreting the overall reasoning process performed by a model. Given a model $f:\mathcal{X}\to\mathcal{Y}$, the interpretation produced by our algorithm is an approximation $T(x)\approx f(x)$, where $T$ is an interpretable model. In this paper, we take $T$ to be a decision tree, which has been established as highly interpretable~\cite{letham2015interpretable,lime,rulelists}. Intuitively, if $T$ is a sufficiently good approximation of $f$, then any issues in $f$ should be reflected in $T$ as well. Thus, the user can understand and debug $f$ by examining $T$; then, the original model $f$ can be deployed so that performance is not sacrificed.
%For example, to understand how $f$ depends on a prejudiced feature, the user can examine nodes in $T$ that branch on that feature, determine how likely it is that test points flow to that node, and estimate the conditional effect size. With this information, the user could carefully train a fairer model for the affected subgroup. Assuming $T$ is a good approximation of $f$, then the dependence of $f$ on the prejudiced feature should be similar.

Previous model extraction approaches have focused on specific model families~\cite{van2007seeing,deng2014interpreting,vandewiele2016genesim}, enabling them to leverage the internal structure of the model. In contrast, our approach is \emph{blackbox}, i.e., it only requires the ability to obtain the output $f(x)\in\mathcal{Y}$ corresponding to a given input $x\in\mathcal{X}$. Thus, our approach works with any model family and is independent of the implementation. Complimentary approaches to interpretability focus on learning interpretable models~\cite{caruana2012intelligible,wang2015falling,slim} or on explaining the model's behavior on specific inputs rather than the model as a whole~\cite{lime}.

The key challenge to learning accurate decision trees is that they often overfit and obtain poor performance, whereas complex models such as random forests and deep neural nets are better regularized~\cite{shallownets}. For example, random forests use ensembles of trees to avoid overfitting to specific features or training points.

Thus, our algorithm uses active learning to construct $T$ from $f$---it actively samples a large number of training points, and computes the corresponding labels $y=f(x)$. The large quantity of data ensures that $T$ does not overfit to the small set of initial training points. We prove that under mild assumptions, by generating a sufficient quantity of data, the extracted tree $T$ converges to the \emph{exact} greedy decision tree, i.e., it avoids overfitting since the sampling error goes to zero.

We implement our algorithm and use it to interpret random forests and neural nets, as well as control policies trained using reinforcement learning. We show that our active learning approach substantially improves over using CART~\cite{CART}, a standard decision tree learning algorithm. Furthermore, we demonstrate how the decision trees extracted can be used to debug issues with these models, for example, to assess the dependence on prejudiced features, to determine why certain models perform worse, and to understand the high-level structure of a learned control policy.

\section{Model Extraction}

We describe our model extraction algorithm.

\subsection{Problem Formulation}

Given a training set $X_{\text{train}}\subseteq\mathcal{X}$ and blackbox access to a function $f:\mathcal{X}\to\mathcal{Y}$, our goal is to learn an interpretable model $T:\mathcal{X}\to\mathcal{Y}$ that approximates $f$. In this paper, we take $T$ to be an axis-aligned decision tree, since these models are both expressive highly interpretable. For simplicity, we focus on the case of classification, i.e., $\mathcal{Y}=[m]=\{1,...,m\}$. We measure performance using accuracy relative to $f$ on a held out test set, i.e.,
$\frac{1}{|X_{\text{test}}|}\sum_{x\in X_{\text{test}}}\mathbb{I}[T(x)=f(x)]$.

\subsection{Algorithm}

Our algorithm is greedy, both for scalability and because it is a natural fit for interpretability, since more relevant features occur higher in the tree.
%Our algorithm first constructs a distribution $\mathcal{P}$ over $\mathcal{X}$ that it uses to guide the model extraction. Using $\mathcal{P}$ in place of the finite training set $X_{\text{train}}$ enables our model extraction algorithm to avoid overfitting. Then, we describe the \emph{exact greedy decision tree} $T^*$, where all decisions are made exactly according to the distribution $\mathcal{P}$ (albeit greedily); note that we cannot construct $T^*$ since we treat $f$ as a blackbox. Finally, we describe how our algorithm estimates $T^*$ using i.i.d. samples from $\mathcal{P}$. In Section~\ref{sec:theory}, we show the estimated tree converges to $T^*$ asymptotically.

\paragraph{Input distribution.}

First, our algorithm constructs a distribution $\mathcal{P}$ over the input space $\mathcal{X}$ by fitting a mixture of axis-aligned Gaussian distributions to the training data using expectation maximization.

\paragraph{Exact greedy decision tree.}

We describe the \emph{exact greedy decision tree} $T^*$. We cannot construct $T^*$ since we treat $f$ as a blackbox; as we describe below, our algorithm approximates $T^*$. Essentially, $T^*$ is constructed greedily as a CART tree~\cite{CART}, except the gain is computed exactly according to $\mathcal{P}$. For example, if the gain is the Gini impurity, then it is computed as follows:
\begin{align*}
\text{Gain}(f,C_N)=1-\sum_{y\in\mathcal{Y}}\text{Pr}_{x\sim\mathcal{P}}[f(x)=y\mid C_N],
\end{align*}
where $C_N$ are the constraints encoding which points flow to the node $N$ in $T^*$ for which a branch is currently being constructed. Similarly, the most optimal leaf labels are computed exactly according to $\mathcal{P}$.

\paragraph{Estimated greedy decision tree.}

Given $n\in\mathbb{N}$, our algorithm estimates $\text{Gain}(f,C_N)$ using $n$ i.i.d. samples $x\sim\mathcal{P}\mid C_N$, where $C_N$ is a conjunction of axis-aligned constraints. We briefly describe how our algorithm obtains such samples. It is straightforward to show that the constraint $C_N$ can be simplified so it contains at most one inequality $(x_i\le t)$ and at most one inequality $(x_i>s)$ per $i\in[d]$. For simplicity, we assume $C_N$ contains both inequalities for each $i\in[d]$:
\begin{align*}
C_N=(s_1\le x_1\le t_1)\wedge...\wedge(s_d\le x_d\le t_d).
\end{align*}
Then, the probability density function of $\mathcal{P}\mid C_N$ is
\begin{align*}
p_{\mathcal{P}\mid C_N}(x)\propto\sum_{j=1}^K\phi_j\prod_{i=1}^dp_{\mathcal{N}(\mu_{ji},\sigma_{ji})\mid(s_i\le x_i\le t_i)}(x_i).
\end{align*}
Since the Gaussians are axis-aligned, the unnormalized probability of each component is
\begin{align*}
\tilde{\phi}_j'
&=\int\phi_j\prod_{i=1}^dp_{\mathcal{N}(\mu_{ji},\sigma_{ji})\mid(s_i\le x_i\le t_i)}(x_i)dx \\
&=\phi_j\prod_{i=1}^d\left(\Phi\left(\frac{t_i-\mu_{ji}}{\sigma_{ji}}\right)-\Phi\left(\frac{s_i-\mu_{ji}}{\sigma_{ji}}\right)\right),
\end{align*}
where $\Phi$ is the cumulative density function of the standard Gaussian distribution $\mathcal{N}(0,1)$. Then, the component probabilities are $\tilde{\phi}=Z^{-1}\tilde{\phi}'$, where $Z=\sum_{j=1}^K\tilde{\phi}_j'$. To sample $x\sim\mathcal{P}\mid C_N$, sample $j\sim\text{Categorical}(\tilde{\phi})$, and
\begin{align*}
x_i&\sim\mathcal{N}(\mu_{ji},\sigma_{ji})\mid(s_i\le x_i\le t_i)\hspace{0.1in}(\text{for each }i\in[d]).
\end{align*}
We use standard algorithms for sampling truncated Gaussian distributions to sample each $x_i$.

\subsection{Theoretical Guarantees}
\label{sec:theory}

The extracted tree $T$ converges to $T^*$ as $n\to\infty$:
\begin{theorem}
\label{thm:exact}
\rm
Assume the exact greedy tree $T^*$ is well defined, and the probability density function $p(x)$ is bounded, continuous, and has bounded support. Then, for any $\epsilon,\delta>0$, there exists $n\in\mathbb{N}$ such that the tree $T$ extracted by our algorithm using $n$ samples satisfies $\text{Pr}_{x\sim\mathcal{P}}[T(x)=T^*(x)]\le\epsilon$, with probability at least $1-\delta$ over the training samples.
\end{theorem}

\section{Evaluation}

We use our model extraction algorithm to interpret several supervised learning models trained on datasets from the UCI Machine Learning Repository~\cite{uci_repository}, as well as a learned control policy from OpenAI Gym~\cite{openai_cartpole}, i.e., the learned control policy $\pi:\mathcal{S}\to\mathcal{A}$.

\subsection{Comparison to CART}

\begin{table*}[t]
\small
\centering
\resizebox{\textwidth}{!}{
\begin{tabular}{llrrl|rrr|rr}
\toprule
\multicolumn{5}{c|}{{\bf Description of Problem Instance}} &
\multicolumn{3}{c|}{{\bf Absolute}} &
\multicolumn{2}{c}{{\bf Relative}} \\
\multicolumn{1}{c}{Dataset} &
\multicolumn{1}{c}{Task} &
\multicolumn{1}{c}{Samples} &
\multicolumn{1}{c}{Features} &
\multicolumn{1}{c|}{Model} &
\multicolumn{1}{c}{$f$} &
\multicolumn{1}{c}{$T$} &
\multicolumn{1}{c|}{$T_{\text{base}}$} &
\multicolumn{1}{c}{$T$} &
\multicolumn{1}{c}{$T_{\text{base}}$} \\
\hline
breast cancer~\cite{wolberg1990multisurface} & classification & 569 & 32 & random forest & 0.966 & {\bf 0.942} & 0.934 & {\bf 0.957} & 0.945 \\
student grade~\cite{cortez2008using} & regression & 382 & 33 & random forest & 4.47 & {\bf 4.70} & 5.10 & {\bf 0.40} & 0.64 \\
wine origin~\cite{forina1991extendible} & classification & 178 & 13 & random forest & 0.981 & {\bf 0.925} & 0.890 & {\bf 0.938} & 0.890 \\
wine origin~\cite{forina1991extendible} & classification & 178 & 13 & neural net & 0.795 & {\bf 0.755} & 0.751 & {\bf 0.913} & 0.905 \\
cartpole~\cite{cartpole_problem} & reinforcement learning & 100 & 4 & control policy & 200.0 & {\bf 190.0} & 35.6 & {\bf 86.8\%} & 83.8\% \\
\bottomrule
\end{tabular}
}
\caption{\small Comparison of the decision tree $T$ extracted by our algorithm to the one $T_{\text{base}}$ extracted by the baseline. We show absolute performance on ground truth and performance relative to the model $f$. For classification (resp., regression), performance is $F_1$ score (resp., MSE) on the test set. For reinforcement learning, it is accuracy on the test set for relative performance, and estimated reward using the decision tree as the policy for absolute performance. We bold the higher score between $T$ and $T_{\text{base}}$.}
\label{tab:results}
\end{table*}

First, we compare our algorithm to a baseline that uses CART to train a decision tree approximating $f$ on the training set $\{(x,f(x))\mid x\in X_{\text{train}}\}$. For both algorithms, we restrict the decision tree to have 31 nodes. We show results in Table~\ref{tab:results}. We show the test set performance of the extracted tree compared to ground truth (or for MDPs, estimated the reward when it is used as a policy), as well as the relative performance compared to the model $f$ on the same test set. Note that our goal is to obtain high relative performance: a better approximation of $f$ is a better interpretation of $f$, even if $f$ has poor performance. Our algorithm outperforms the baseline on every problem instance.

\subsection{Examples of Use Cases}

We show how the extracted decision trees can be used to interpret and debug models.

\paragraph{Use of invalid features.}

Using an invalid feature is a common problem when training models. In particular, some datasets contain multiple response variables; then, one response should not be used to predict the other. For example, the breast cancer dataset contains two response variables indicating cancer recurrence: the length of time before recurrence and whether recurrence occurs within 24 months. This issue can be thought of as a special case of using a non-causal feature, an important problem in healthcare settings. We train a random forest $f$ to predict whether recurrence occurs within 24 months, where time to recurrence is incorrectly included as a feature. Then, we extract a decision tree approximating $f$ of size $k=7$ nodes, using 10 random splits of the dataset into training and test sets. The invalid feature occured in every extracted tree, and as the top branch in 6 of the 10 trees.

\paragraph{Use of prejudiced features.}

We can use our algorithm to evaluate how a model $f$ depends on prejudiced features. For example, gender is a feature in the student grade dataset, and may be considered sensitive when estimating student performance. However, if we simply omit gender, then $f$ may reconstruct it from the remaining features. For a model $f$ trained with gender available, we show how a decision tree extracted from $f$ can be used to understand how $f$ depends on gender. Our approach does not guarantee fairness, but it can be useful for evaluating the fairness of $f$.

We extract decision trees $T$ from the random forests $f$ trained on 10 random splits of the student grades dataset. The top features are consistently grades in other classes or number of failing grades received in the past. Gender occurs below these features (at the fourth or fifth level) in 7 of 10 of the trees. We can estimate the overall effect of changing the gender label:
\begin{align*}
\Delta=\mathbb{E}_{x\sim\mathcal{P}}[f(x)\mid\text{male}]-\mathbb{E}_{x\sim\mathcal{P}}[f(x)\mid\text{female}].
\end{align*}
When gender occurs, $\Delta$ is between 0.31 and 0.70 grade points (average 0.49) out of 20 total grade points. For the remaining models, $\Delta$ is between 0.11 and 0.32 (average 0.25). Thus, the extracted tree includes gender when $f$ has a relatively large dependence on gender.

Furthermore, because $T$ approximates $f$, we can use it to identify a subgroup of students where $f$ has particularly strong dependence on gender. In particular, points that flow to the internal node $N$ of $T$ branching on gender are a subset of inputs whose label $T(x)\in\mathcal{Y}$ is determined in part by gender. We can use $T$ to measure the dependence on gender within this subset:
\begin{align*}
\Delta_N=\mathbb{E}_{x\sim\mathcal{P}}[f(x)\mid C_{N_L}]-\mathbb{E}_{x\sim\mathcal{P}}[f(x)\mid C_{N_R}],
\end{align*}
where $N_L$ and $N_R$ are the left and right children of $N$.

Also, we can estimate the fraction of students in this subset using the test set, i.e., $P=\sum_{x\in X_{\text{test}}}\mathbb{I}[x\in\mathcal{F}(C_N)]$. Finally, $P\cdot\Delta_N/\Delta$ measures the fraction of the overall dependence of $f$ on gender that is accounted for by the subtree rooted at $N$. For models where gender occurs in the extracted tree, the subgroup effect size $\Delta_N$ ranged from 0.44 to 0.77 grade points, and the estimated fraction of students in this subroup ranged from 18.3\% to 39.1\%. The two trees that had the largest effect size had $\Delta_N$ of 0.77 and 0.43, resp., and $P$ of 39.1\% and 35.7\%, resp. The identified subgroup accounted for 67.3\% and 65.6\% of the total effect of gender, resp.

Having identified a subgroup of students likely to be adversely affected, the user might be able to train a better model specifically for this subgroup. In 5 of the 7 extracted trees where gender occurs, the affected students were students with low grades, in particular, the 27\% of students who scored fewer than 8.5 points in another class. This fine-grained understanding of $f$ relies on the extracted model, and cannot be obtained using feature importance metrics alone.

\paragraph{Comparing models.}

We can use the extracted decision trees to compare different models trained on the same dataset, and gain insight into why some models perform better than others. For example, random forests trained on the wine origin dataset performed very well, all achieving an $F_1$ score of at least 0.961. In contrast, the performance of the neural nets was bimodal---5 had $F_1$ score of at least 0.955, and the remaining had an $F_1$ score of at most 0.741.

We examined the top 3 layers of the extracted decision trees $T$, and made two observations. First, occurrence of the feature ``chlorides'' in $T$ was almost perfectly correlated with poor performance of the neural nets. This feature occured in only one of the 10 trees extracted from random forests, and in none of the trees extracted from high performing neural nets. A weaker observation was the branching of $T$ on the feature ``alcohol'', which is a very important feature---it is the top branch for all but one of the 20 extracted decision trees. For the high performing models, the branch threshold $t$ tended to be higher (749.8 to 999.6) than those for the poorly performing models (574.4 to 837.3). The latter observation relies on having an extracted model---feature influence metrics alone are insufficient.

\paragraph{Understanding control policies.}

We can use the extracted decision tree to understand a control policy. For example, we extracted a decision tree of size $k=7$ from the cartpole control policy. While its estimated reward of 152.3 is lower than for larger trees, it captures a significant fraction of the policy behavior. The tree says to move the cart to the right exactly when
\begin{align*}
(\text{pole velocity}\ge-0.286)\wedge(\text{pole angle}\ge-0.071),
\end{align*}
where the pole velocity is in $[-2.0,2.0]$ and the pole angle is in $[-0.5,0.5]$. In other words, move the cart to the right exactly when the pole is already on the right relative to the cart, and the pole is also moving toward the left (or more precisely, not moving fast enough toward the right). This policy is asymmetric, focusing on states where the cart is moving to the left. Examining an animation of simulation, this bias occurs because the cart initially moves toward the left, so the portion of the state space where the cart is moving toward the right is relatively unexplored.

\section{Conclusions}

We have proposed model extraction as an approach for interpreting blackbox models, and shown how it can be used to interpret a variety of different kinds of models. Important directions for future work include devising algorithms for model extraction using more expressive input distributions, and developing new ways to gain insight from the extracted decision trees.

\newpage

\bibliographystyle{plain}
\bibliography{paper}

\end{document}